\documentclass{article} 
\PassOptionsToPackage{table}{xcolor}
\usepackage{iclr2024_conference,times}

\usepackage{amsmath,amsfonts,bm}

\def\eqref#1{equation~\ref{#1}}

\def\1{\bm{1}}

\DeclareMathAlphabet{\mathsfit}{\encodingdefault}{\sfdefault}{m}{sl}
\SetMathAlphabet{\mathsfit}{bold}{\encodingdefault}{\sfdefault}{bx}{n}

\usepackage{xspace}
\usepackage{hyperref}
\usepackage{url}
\usepackage{times}
\usepackage{epsfig}
\usepackage{subfigure}
\usepackage{graphicx}
\usepackage{amsmath}
\usepackage{amssymb}
\usepackage{booktabs}
\usepackage{float}
\usepackage{nicefrac}
\usepackage{boldline}
\usepackage{multirow}
\usepackage{bm}
\usepackage{color}
\usepackage{arydshln}
\usepackage[accsupp]{axessibility}
\usepackage{caption}
\usepackage{cleveref}

\title{PanoDiffusion: 360-degree Panorama Outpainting via Diffusion}

\author{Tianhao Wu$^1$, Chuanxia Zheng$^2$ \& Tat-Jen Cham$^1$ \\
$^1$Nanyang Technological University \\
\texttt{tianhao001@e.ntu.edu.sg, astjcham@ntu.edu.sg} \\
$^2$University of Oxford \\
\texttt{cxzheng@robots.ox.ac.uk}
}

\newcommand{\mname}{PanoDiffusion\xspace}

\begin{document}

\maketitle

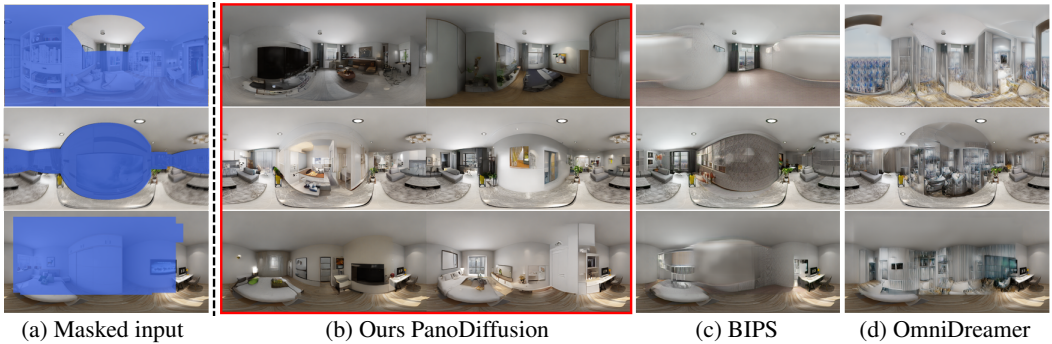
\begin{figure}[b!]
            \vspace{-0.3cm}
		\centering
		\hsize=\textwidth
		\includegraphics[width=\textwidth]{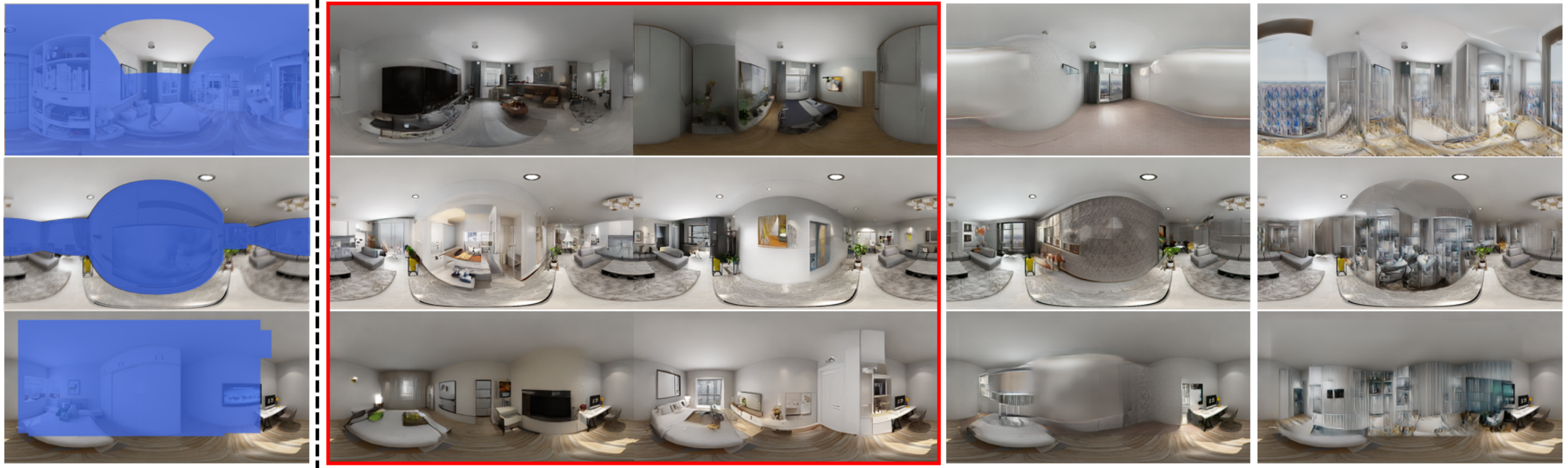}
        \begin{picture}(0,0)
            \put(-190,3){\footnotesize (a) Masked input}
            \put(-75,3){\footnotesize (b) Ours \mname}
            \put(65,3){\footnotesize (c) BIPS}
            \put(127,3){\footnotesize (d) OmniDreamer}
        \end{picture}
        \vspace{-0.0cm}
        \caption{\textbf{Example results of 360$^\circ$ Panorama Outpainting on various masks.} Compared to BIPS \citep{oh2022bips} and OmniDreamer \citep{akimoto2022diverse}, our model not only effectively generates semantically meaningful content and plausible appearances with many objects, such as beds, sofas and TV's, 
        but also provides \emph{multiple} and \emph{diverse} solutions for this ill-posed problem. 
        (Masked regions are shown in blue for better visualization.
        \textcolor[RGB]{255,0,0}{Zoom in to see the details.})}
        \label{fig: outpainting examples}
\end{figure}

\begin{abstract}
Generating complete 360\textdegree{} panoramas from narrow field of view images is ongoing research as omnidirectional RGB data is not readily available. 
Existing GAN-based approaches face some barriers to achieving higher quality output, and have poor generalization performance over different mask types. In this paper, we present our 360\textdegree{} indoor RGB-D panorama outpainting model using latent diffusion models (LDM), called \mname. 
We introduce a new bi-modal latent diffusion structure that utilizes both RGB and depth panoramic data during training, which works surprisingly well to outpaint \emph{depth-free} RGB images during inference. 
We further propose a novel technique of introducing progressive camera rotations during each diffusion denoising step, which leads to substantial improvement in achieving panorama wraparound consistency. 
Results show that our \mname not only significantly outperforms state-of-the-art methods on RGB-D panorama outpainting by producing diverse well-structured results for different types of masks, but can also synthesize high-quality depth panoramas to provide realistic 3D indoor models.
\end{abstract}

\section{Introduction}

Omnidirectional 360\textdegree{} panoramas serve as invaluable assets in various applications, such as lighting estimation \citep{gardner2017learning,gardner2019deep,song2019neural} and new scene synthesis \citep{somanath2021hdr} in the Augmented and Virtual Reality (AR \& VR). 
However, an obvious limitation is that capturing, collecting, and restoring a dataset with 360\textdegree{} images is a \emph{high-effort} and \emph{high-cost} undertaking \citep{akimoto2019360,akimoto2022diverse}, while manually creating a 3D space from scratch can be a demanding task \citep{lee2017joint, choi2015robust,newcombe2011kinectfusion}.

To mitigate this dataset issue, the latest learning methods \citep{akimoto2019360,somanath2021hdr,akimoto2022diverse,oh2022bips} have been proposed, with a specific focus on \emph{generating omnidirectional RGB panoramas from narrow field of view (NFoV) images.} 
These methods are typically built upon Generative Adversarial Networks (GANs) \citep{goodfellow2014generative}, which have shown remarkable success in creating new content. 
However, GAN architectures face some notable problems, including 
\textbf{1)} mode collapse (seen in Fig.~\ref{fig: outpainting examples}(c)), 
\textbf{2)} unstable training \citep{salimans2016improved}, and 
\textbf{3)} difficulty in generating multiple structurally reasonable objects \citep{epstein2022blobgan}.
These limitations lead to obvious artifacts in synthesizing complex scenes (Fig.~\ref{fig: outpainting examples}). 


The recent endeavors of \citep{lugmayr2022repaint,li2022sdm,xie2023smartbrush, wang2023imagen} directly adopt the latest latent diffusion models (LDMs) \citep{rombach2022high} in image inpainting tasks, 
which achieve a stable training of generative models and spatially consistent images. 
However, specifically for a 360\textdegree{} panorama outpainting scenario, these inpainting works usually lead to grossly distorted results. 
This is because: 
\textbf{1)} the missing (masked) regions in 360\textdegree{} panorama outpainting is generally \emph{much larger} than masks in normal inpainting and
\textbf{2)}  it necessitates generating \emph{semantically reasonable objects} within a given scene, as opposed to merely filling in generic background textures in an empty room (as shown in Fig.~\ref{fig: outpainting examples} (c)). 
To achieve this, we propose an alternative method for 360\textdegree{} indoor panorama outpainting via the latest latent diffusion models (LDMs) \citep{rombach2022high}, termed as \mname.
Unlike existing diffusion-based inpainting methods, we introduce \emph{depth} information through a novel \emph{bi-modal} latent diffusion structure during the \emph{training}, 
which is also significantly different from the latest concurrent works \citep{tang2023MVDiffusion,lu2023autoregressive} that aims for \emph{text-guided} 360\textdegree{} panorama image generation.
Our \emph{key motivation} for doing so is that the depth information is crucial for helping the network understand the physical structure of objects and the layout of the scene \citep{ren2012rgb}.
It is worth noting that our model only uses partially visible RGB images as input during \emph{inference}, \emph{without} requirement for any depth information, yet achieving significant improvement on both RGB and depth synthesis (Tables \ref{tab:rgb_quant} and \ref{tab:ablations}).



Another distinctive challenge in this task stems from the unique characteristic of panorama images: 
\textbf{3)} both ends of the image must be aligned to ensure the integrity and \emph{wraparound consistency} of the entire space, given that the indoor space lacks a definitive starting and ending point.
To enhance this property in the generated results, we introduce two strategies:
First, during the \emph{training} process, a \emph{camera-rotation} approach is employed to randomly crop and stitch the images for data augmentation. 
It encourages the networks to capture information from different views in a 360\textdegree{} panorama. 
Second, a \emph{two-end alignment} mechanism is applied at each step of the \emph{inference} denoising process (Fig.~\ref{fig:two-end align}), which explicitly enforces the two ends of an image to be wraparound-consistent.

We evaluate the proposed method on the Structured3D dataset \citep{zheng2020structured3d}. Experimental results demonstrate that our \mname not only significantly outperforms previous state-of-the-art methods, but is also able to provide \emph{multiple} and \emph{diverse} well-structured results for different types of masks (Fig.~\ref{fig: outpainting examples}). In summary, our main contributions are as follows:
\begin{itemize}
    \item A new bi-modal latent diffusion structure that utilizes both RGB and depth panoramic data to better learn spatial layouts and patterns during training, but works surprisingly well to outpaint normal RGB-D panoramas during inference, \emph{even without depth input};
    \item A novel technique of introducing progressive camera rotations during \emph{each} diffusion denoising step, which leads to substantial improvement in achieving panorama wraparound consistency;
    \item With either partially or fully visible RGB inputs, our \mname can synthesize high-quality indoor RGB-D panoramas simultaneously to provide realistic 3D indoor models.
\end{itemize}

\section{Related Work}

\subsection{Image Inpainting/Outpainting}

Driven by advances in various generative models, such as VAEs \citep{kingma2013auto} and GANs \citep{goodfellow2014generative},
a series of learning-based approaches \citep{pathak2016context,iizuka2017globally,yu2018generative,zheng2019pluralistic,zhao2020large,Wan_2021_ICCV,zheng2022bridging} have been proposed to generate semantically meaningful content from a partially visible image.
More recently, state-of-the-art methods \citep{lugmayr2022repaint,li2022sdm,xie2023smartbrush,wang2023imagen} directly adopt the popular diffusion models \citep{rombach2022high} for image inpainting, achieving high-quality completed images with consistent structure and diverse content.
However, these diffusion-based models either focus on background inpainting, or require input text as guidance to produce plausible objects within the missing regions.
This points to an existing gap in achieving comprehensive and contextually rich inpainting/outpainting results across a wider spectrum of scenarios, especially in the large scale 360\textdegree{} Panorama scenes.



\subsection{360\textdegree{} Panorama Outpainting}
\vspace{-0.2cm}
Unlike NFoV images, 360\textdegree{} panorama images are subjected to nonlinear perspective distortion, such as \emph{equirectangular projection}. 
Consequently, objects and layouts within these images appear substantially distorted, particularly those closer to the top and bottom poles. 
The image completion has to not only preserve the distorted structure but also ensure visual plausibility, with the additional requirement of \emph{wraparound-consistency at both ends}.
Previous endeavors \citep{akimoto2019360,somanath2021hdr} mainly focused on deterministic completion of 360\textdegree{} RGB images, with BIPS \citep{oh2022bips} further extending this to RGB-D panorama synthesis. 
In order to generate diverse results \citep{zheng2019pluralistic,zheng2021pluralistic}, various strategies have been employed. 
For instance, SIG-SS \citep{hara2021spherical} leverages a symmetry-informed CVAE, while OmniDreamer \citep{akimoto2022diverse} employs transformer-based sampling. 
In contrast, our \mname is built upon DDPM, wherein each reverse diffusion step inherently introduces stochastic, naturally resulting in \emph{multiple} and \emph{diverse} results.
Concurrently with our work, MVDiffusion \citep{tang2023MVDiffusion} generates panorama images by sampling consistent multi-view images, and AOGNet \citep{lu2023autoregressive} does 360\textdegree{} outpainting through an autoregressive process. 
Compared to the concurrent models, our \mname excels in generating semantically multi-objects for large masked regions, \emph{without the need of text prompts.}
More importantly, \mname is capable of simultaneously generating the corresponding RGB-D output, using only partially visible RGB images as input during the \emph{inference}.  


\section{Methods}
\vspace{-0.2cm}
\begin{figure*}[tb!]
    \centering
    \includegraphics[width=\linewidth]{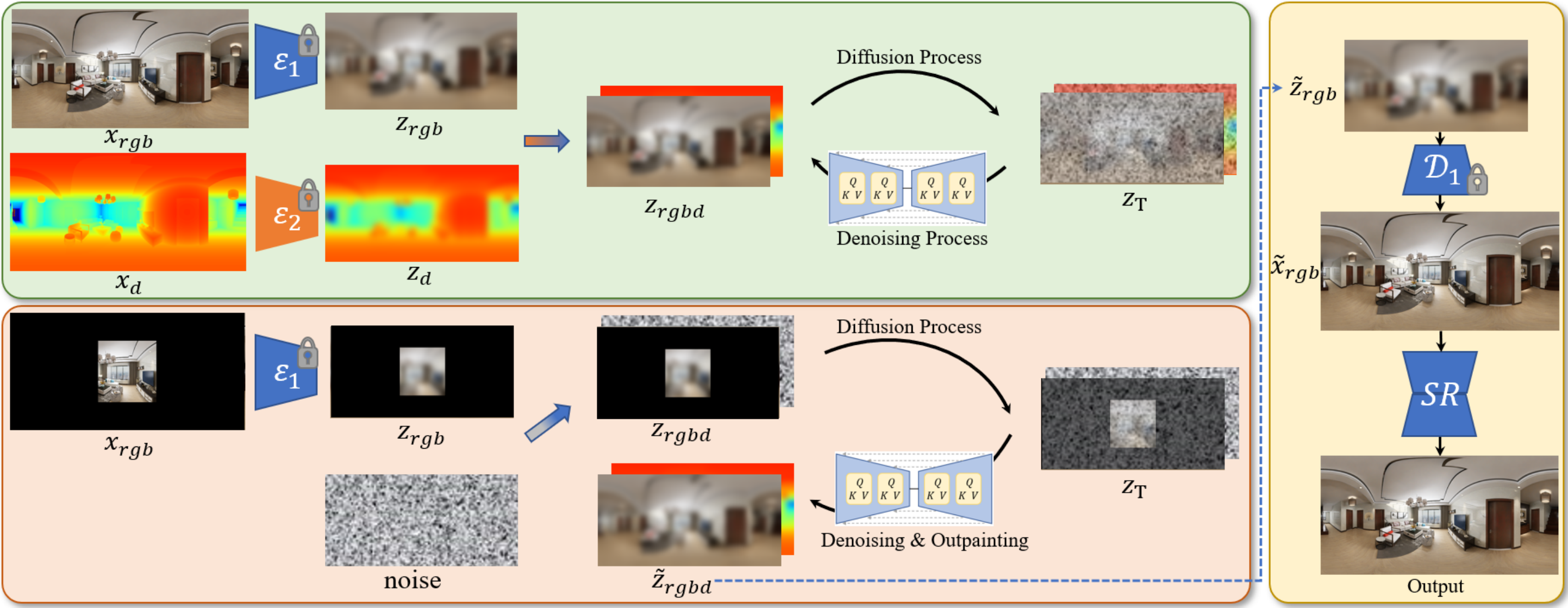}
    \begin{picture}(0,0)
    \put(60,158){\scriptsize (a) Training Stage}
    \put(60, 82){\scriptsize (b) Inference Stage}
    \put(128, 158){\scriptsize (c) Refine Stage}
    \end{picture}
    \vspace{-0.6cm}
   \caption{\textbf{The overall pipeline of our proposed \mname method.} 
   (a) During training, the model is optimized for RGB-D panorama synthesis, without the mask. 
   (b) During inference, however, the depth information is \emph{no longer needed} for masked panorama outpainting. 
   (c) Finally, a super-resolution model is implemented to further enhance the high-resolution outpainting. 
   We only show the input/output of each stage and omit the details of circular shift and adding noise.
   Note that the VQ-based encoder-decoders are pre-trained in advance, and fixed in the rest of our framework. 
   }
   \vspace{-0.3cm}
\label{fig: overview}
\end{figure*}
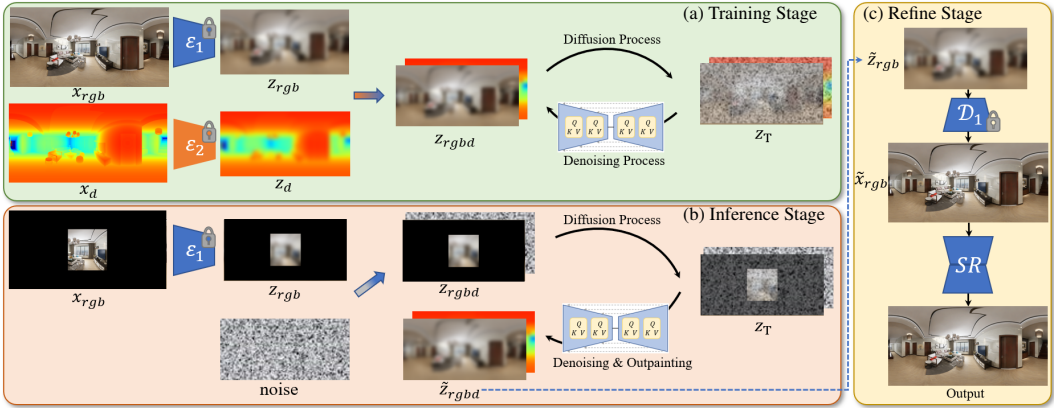

Given a 360\textdegree{} image $x\in\mathbb{R}^{H\times W\times C}$, degraded by a number of missing pixels to become a masked image $x_m$, our main goal is to infer semantically meaningful content with reasonable geometry for the missing regions, while simultaneously generating visually realistic appearances. 
While this task is conceptually similar to conventional learning-based image inpainting, it presents greater challenges due to the following differences: 
\textbf{1)} our \textbf{output} is a \emph{360\textdegree{} RGB-D panorama that requires wraparound consistency}; 
\textbf{2)} the \textbf{masked/missing} areas are generally \emph{much larger} than the masks in traditional inpainting; 
\textbf{3)} our \textbf{goal} is to \emph{generate multiple appropriate objects} within a scene, instead of simply filling in with the generic background; 
\textbf{4)} the completed results should be structurally plausible, which can be reflected by a reasonable depth map. 

To tackle these challenges, we propose a novel diffusion-based framework for 360\textdegree{} panoramic outpainting, called \mname.
The training process, as illustrated in Fig.~\ref{fig: overview}(a), starts with two branches dedicated to RGB $x$ and depth $d_x$ information. 
Within each branch, following \citep{rombach2022high}, the input data is first embedded into the latent space using the corresponding pre-trained VQ model.
These representations are then concatenated to yield $z_{rgbd}$, which subsequently undergoes the forward diffusion step to obtain $z_T$.
The resulting $z_T$ is then subjected to inverse denoising, facilitated by a trained UNet, ultimately returning to the original latent domain.
Finally, the pre-trained decoder is employed to rebuild the completed RGB-D results.

During inference, our system takes a masked RGB image as input and conducts panoramic outpainting. 
It is noteworthy that our proposed model does \emph{not} inherently require harder-to-acquire depth maps as input, \emph{relying solely on a partial RGB image} (Fig.~\ref{fig: overview}(b)). 
The output is further super-resolved into the final image in a refinement stage (Fig.~\ref{fig: overview}(c)).


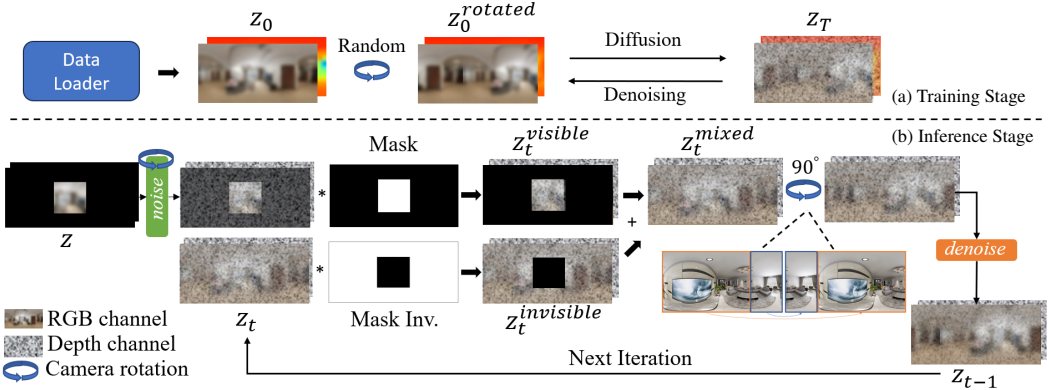
\begin{figure*}[tb!]
    \centering
    \includegraphics[width=\linewidth]{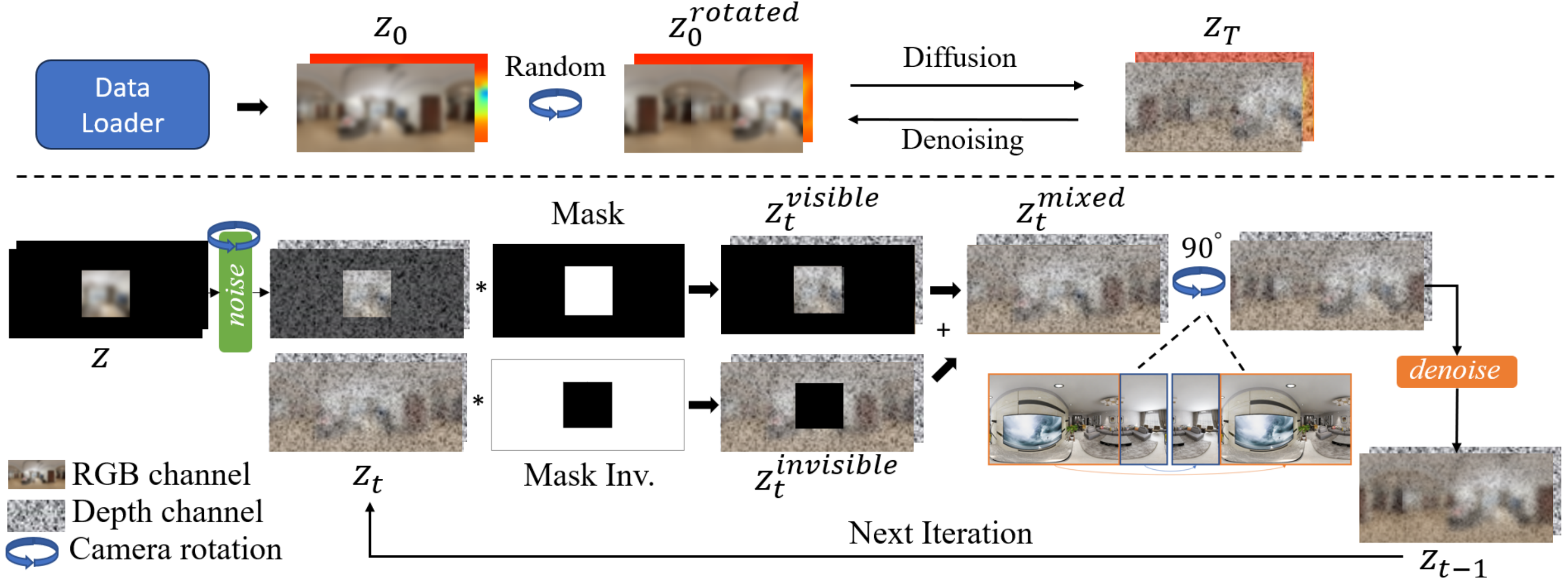}
    \begin{picture}(0,0)
    \put(140,120){\scriptsize (a) Training Stage}
    \put(140, 105){\scriptsize (b) Inference Stage}
    \end{picture}
    \vspace{-0.6cm}
   \caption{\textbf{Our LDM outpainting structure with camera rotation mechanism.} During training (a), we \textbf{randomly} select a rotation angle to generate a new panorama for data augmentation. During inference (b), we sample the visible region from the encoded features (above) and the invisible part from the denoising output (below). The depth map is \emph{not needed}, and is set to random noise. At each denoising step, we crop a \textbf{90\textdegree{}}-equivalent area of the intermediate result from the right and stitch it to the left, denoted by the circle following $z_{t}^{mixed}$ ---  
   this strongly improves wraparound consistency.
   }
\label{fig: repaint}
\end{figure*}


\subsection{Preliminaries}
\vspace{-6pt}
\paragraph{Latent Diffusion.} Our \mname builds upon the latest Latent Diffusion Model (LDM)~\citep{rombach2022high}, which executes the denoising process in the latent space of an autoencoder. 
This design choice yields a twofold advantage: it reduces computational costs while maintaining a high level of visual quality by storing the domain information in the encoder $\mathcal{E}(\cdot)$ and decoder $\mathcal{D}(\cdot)$. 
During the training, the given image $x_0$ is initially embedded to yield the latent representation $z_0=\mathcal{E}(x_0)$, which is then perturbed by adding the noise in a Markovian manner:
\begin{equation}\label{eq:ldmforward}
  q(z_t|z_{t-1}) = \mathcal{N}(z_t;\sqrt{1-\beta_t}z_{t-1},\beta_t I),
\end{equation}
where $t=[1,\cdots, T]$ is the number of steps in the forward process.
The hyperparameters $\beta_t$ denote the noise level at each step $t$.
For the denoising process, the network in LDM is trained to predict the noise as proposed in DDPM~\citep{ho2020denoising}, 
where the training objective can be expressed as:
\begin{equation}\label{eq:ldmbackward}
   \mathcal{L} = \mathbb{E}_{\mathcal{E}(x_0),\epsilon\thicksim\mathcal{N}(0,I), t}
   [||\epsilon_\theta(z_t, t)-\epsilon||_2^2]
\end{equation}

\vspace{-12pt}
\paragraph{Diffusion Outpainting.}


The existing pixel-level diffusion inpainting methods~\citep{lugmayr2022repaint,horita2022structure} are conceptually similar to that used for image generation, except  $x_t$ \emph{incorporates partially visible information}, rather than purely sampling from a Gaussian distribution during the inference.
In particular, let $x_0$ denote the original image in step 0, while $x_0^{visible}=m\odot{x_0}$ and $x_0^{invisible}=(1-m)\odot{x_0}$ contain the visible and missing pixels, respectively. 
Then, as shown in Fig.~\ref{fig: repaint}, the reverse denoising sampling process unfolds as follows:
\begin{gather}\label{eq:diff}
    x^{visible}_{t}\sim{}q(x_t|x_{t-1}), \\
    x^{invisible}_{t-1}\sim{}p_\theta(x_{t-1}|x_t), \\
    x_{t-1}=m\odot{}x^{visible}_{t-1}+(1-m)\odot{}x^{invisible}_{t-1}.
\end{gather}
Here, $q$ is the forward distribution in the diffusion process and $p_\theta$ is the inverse distribution. After $T$ iterations, $x_0$ is restored to the original image space.

\vspace{-8pt}
\paragraph{Relation to Prior Work.} In contrast to these inpainting methods at pixel-level, our \mname builds upon the LDM. Despite the fact that the original LDM provided the ability to inpainting images, such inpainting focuses on removing objects from the image, rather than generating a variety of meaningful objects in panoramic outpainting.
In short, the $x_0$ is embedded into the latent space, yielding $z_0=\mathcal{E}(x_0)$, while the subsequent sampling process follows the equations (3)-(5).
The \emph{key motivation} behind this is to perform our task on higher resolution 512$\times$1024 panoramas.
More importantly, we opt to go beyond RGB outpainting, and to deal with RGB-D synthesis (Sec.~\ref{Sec: Bi-modal Latent Diffusion Model}), which is useful for downstream tasks in 3D reconstruction.
Additionally, existing approaches can \emph{not} ensure the \emph{wraparound consistency}  during completion, while our proposed \emph{rotational outpainting mechanism} in Sec.~\ref{Sec: Camera Rotation} significantly improves such a wraparound consistency.


\begin{figure*}[tb!]
    \centering
    \includegraphics[width=\linewidth]{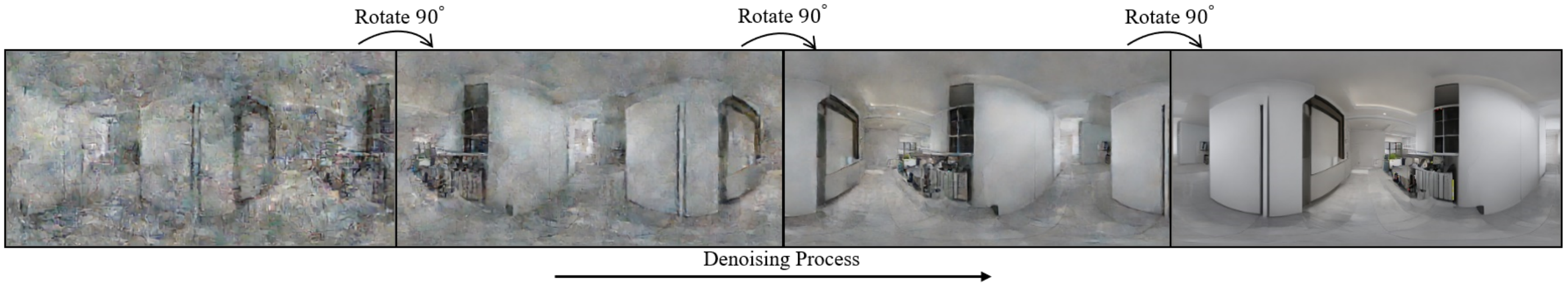}
    \vspace{-0.6cm}
    \caption{\textbf{An example of our two-end alignment mechanism.} During inference, we rotate the scene for 90\textdegree{} in \emph{each} denoising step. Within a total of 200 sampling steps,  our \mname will effectively achieve wraparound consistency.}
    \label{fig:two-end align}
    \vspace{-0.3cm}
\end{figure*}

\subsection{Wraparound Consistency Mechanism}
\label{Sec: Camera Rotation}
\vspace{-6pt}
\paragraph{Camera Rotated Data Augmentation.}
It is expected that the two ends of any 360\textdegree{} panorama should be seamlessly aligned, creating a consistent transition from one end to the other.
This is especially crucial in applications where a smooth visual experience is required, such as 3D reconstruction and rendering.
To promote this property, we implement a \emph{circular shift} data augmentation, termed \emph{camera-rotation}, to train our \mname.
As shown in Fig.~\ref{fig: repaint}(a), we randomly select a rotation angle, which is subsequently employed to crop and reassemble image patches, generating a new panorama for training purposes.

\vspace{-8pt}
\paragraph{Two-Ends Alignment Sampling.}
While the above \emph{camera-rotation} technique can improve the model's implicit grasp of the wraparound consistency using the augmentation of data examples,
it may \emph{not} impose strong enough constraints on wraparound alignment of the results. 
Therefore, in the inference process, we introduce a \emph{novel two-end alignment mechanism} that can be seamlessly integrated into our latent diffusion outpainting process.
In particular, the reverse denoising process within the DDPM is characterized by multiple iterations, rather than a single step.
During \emph{each iteration}, we apply the camera-rotation operation, entailing 90\textdegree{} rotation of both the latent vectors and mask, before performing a denoising outpainting step.
This procedure more effectively connects the two ends of the panorama from the previous step, 
resulting in significant improvement in visual results (as shown in Fig.~\ref{fig: two-end align examples}). 
Without changing the size of the images, generating overlapping content, or introducing extra loss functions, 
we provide `hints' to the model by rotating the panorama horizontally, thus enhancing the effect of alignment at both ends (examples shown in Fig.~\ref{fig:two-end align}).

\subsection{Bi-modal Latent Diffusion Model}
\label{Sec: Bi-modal Latent Diffusion Model}
In order to deal with RGB-D synthesis, one straightforward idea could be to use Depth as an explicit condition during training and inference, where the depth information may be compressed into latent space and then introduced into the denoising process of the RGB images via concatenation or cross-attention. 
However, we found that such an approach often leads to blurry results in our experiments.
Alternatively, using two parallel LDMs to reconstruct Depth and RGB images separately, together with a joint loss, may also appear to be an intuitive solution.
Nonetheless, this idea presents significant implementation challenges due to the computational resources required for multiple LDMs. 

Therefore, we devised a bi-modal latent diffusion structure to introduce depth information while generating high-quality RGB output.
It is important to note that this depth information is \emph{solely necessary during the training phase.}
Specifically, we trained two VAE models independently for RGB and depth images, and then concatenate $z_{rgb}\in{\mathbb{R}^{h\times{}w\times{}3}}$ with $z_{depth}\in{\mathbb{R}^{h\times{}w\times{}1}}$ at the latent level to get $z_{rgbd}\in{\mathbb{R}^{h\times{}w\times{}4}}$.
The training of VAEs is exactly the same as in \citep{rombach2022high} with downsampling factor $f$=$4$. 
Then we follow the standard process to train an unconditional DDPM with $z_{rgbd}$ via a variant of the original LDM loss:

\vspace{-16pt}
\begin{equation}
\begin{split}
\label{eq: LDM loss}
\mathcal{L}_{RGB-D} =\mathbb{E}_{z_{rgbd}, \epsilon\sim{}\mathcal{N}(0,1),t}[\lVert\epsilon_{\theta}(z_t,t)-\epsilon\rVert_2^2],
z_{rgbd}=\mathcal{E}_1(x)\oplus\mathcal{E}_2(d_x)
\end{split}
\end{equation}

\vspace{-6pt}
Reconstructed RGB-D images can be obtained by decoupling $z_{rgbd}$ and decoding. 
It is important to note that during training, we use the full RGB-D image as input, \emph{without masks}. 
Conversely, during the inference stage, the model can perform outpainting of the masked RGB image directly \emph{without any depth input}, with the fourth channel of $z_{rgbd}$ replaced by random noise.

\subsection{RefineNet} 
\label{sec: super resolution}
Although mapping images to a smaller latent space via an autoencoder prior to diffusion can save training space and thus allow larger size inputs, the panorama size of 512$\times$1024 is still a heavy burden for LDM~\citep{rombach2022high}. 
Therefore, we adopt a two-stage approach to complete the outpainting task. 
Initially, the original input is downscaled to 256$\times$512 as the input to the LDM.
Correspondingly, the image size of the LDM output is also 256$\times$512. 
Therefore, an additional module is needed to upscale the output image size to 512$\times$1024. 
Since panorama images are distorted and the objects and layouts do not follow the regular image patterns, we trained a super-resolution GAN model for panoramas to produce visually plausible results at a higher resolution.

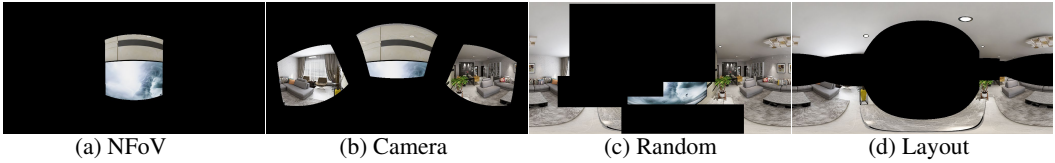
\begin{figure}[tb!]
    \centering
    \includegraphics[width=\linewidth]{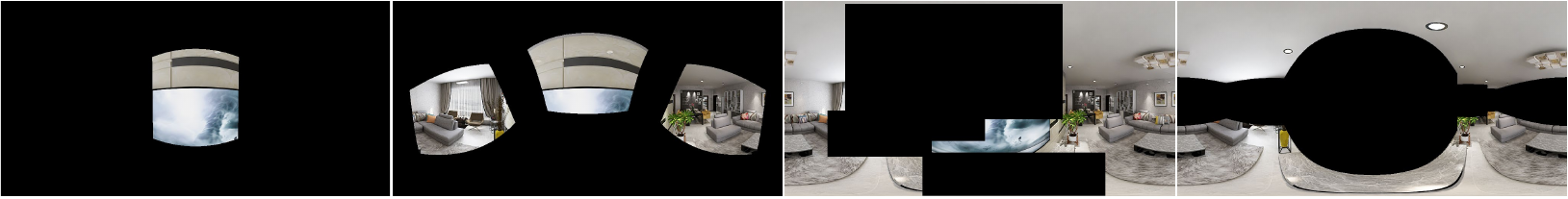}
    \begin{picture}(0,0)
        \put(-170,4){\footnotesize (a) NFoV}
        \put(-70,4){\footnotesize (b) Camera}
        \put(30,4){\footnotesize (c) Random}
        \put(130,4){\footnotesize (d) Layout}
    \end{picture}
    \vspace{-0.3cm}
    \caption{\textbf{Examples of various mask types.} See text for details.}
    \label{fig: mask example}
    \vspace{-0.3cm}
\end{figure}

\section{Experiments}

\subsection{Experimental Details}
\vspace{-6pt}
\paragraph{Dataset.} We estimated our model on the Structured3D dataset \citep{zheng2020structured3d}, which provides 360\textdegree{} indoor RGB-D data following equirectangular projection with a 512$\times$1024 resolution. We split the dataset into 16930 train, 2116 validation, and 2117 test instances.

\vspace{-8pt}
\paragraph{Metrics.} For RGB outpainting, due to large masks, we should not require the completed image to be exactly the same as the original image, since there are many plausible solutions (e.g. new furniture and ornaments, and their placement). Therefore, we mainly report the following dataset-level metrics: 1) Fr$\acute{e}$chet Inception Distance (FID) \citep{heusel2017gans}, 2) Spatial FID (sFID) \citep{nash2021generating}, 3) density and coverage \citep{naeem2020reliable}. 
FID compares the distance between distributions of generated and original images in a deep feature domain, while sFID is a variant of FID that uses spatial features rather than the standard pooled features. 
Additionally, density reflects how accurate the generated data is to the real data stream, while coverage reflects how well the generated data generalizes the real data stream. 
For depth synthesis, we use RMSE, MAE, AbsREL, and Delta1.25 as implemented in \citep{cheng2018depth,zheng2018t2net}, which are commonly used to measure the accuracy of depth estimates.

\vspace{-8pt}
\paragraph{Mask Types.}
Most works focused on generating omnidirectional images from NFoV images (Fig. \ref{fig: mask example}(a)). 
However, partial observability may also occur due to sensor damage in 360\textdegree{} cameras. 
Such masks can be roughly simulated by randomly sampling a number of NFoV camera views within the panorama (Fig.~\ref{fig: mask example}(b)). 
We also experimented with other types of masks, such as randomly generated regular masks (Fig.~\ref{fig: mask example}(c)). 
Finally, the regions with floors and ceilings in panoramic images are often less interesting than the central regions. 
Hence, we also generated layout masks that muffle all areas except floors and ceilings, to more incisively test the model's generative power (Fig.~\ref{fig: mask example}(d)).

\vspace{-8pt}
\paragraph{Baseline Models.} For RGB panorama outpainting, we mainly compared with the following state-of-the-art methods: including image inpainting models LaMa \citep{suvorov2022resolution}$_{\text{\scriptsize{WACV'2022}}}$ and TFill \citep{zheng2022bridging}$_{\text{\scriptsize{CVPR'2022}}}$, panorama outpainting models BIPS \citep{oh2022bips}$_{\text{\scriptsize{ECCV'2022}}}$ and OmniDreamer \citep{akimoto2022diverse}$_{\text{\scriptsize{CVPR'2022}}}$, diffusion-based image inpainting models Repaint \citep{lugmayr2022repaint}$_{\text{\scriptsize{CVPR'2022}}}$ and Inpaint Anything \citep{yu2023inpaint}$_{\text{\scriptsize{arXiv'2023}}}$. 
To evaluate the quality of depth panorama, we compare our method with three image-guided depth synthesis methods including BIPS \citep{oh2022bips}, NLSPN \citep{park2020non}, and CSPN \citep{cheng2018depth}.
All models are retrained on the Structured3D dataset using their publicly available codes.

\subsection{Main Results}

\begin{figure*}[tb!]
    \centering
    \includegraphics[width=\linewidth]{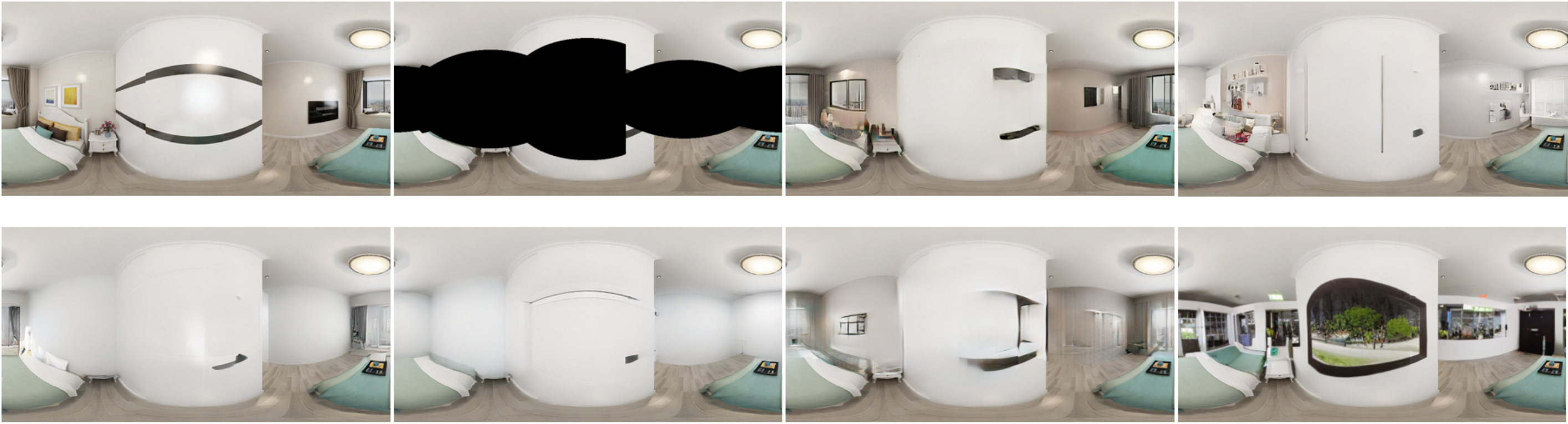}
    \begin{picture}(0,0)
        \put(-178,62){\footnotesize (a) Ground Truth}
        \put(-80,62){\footnotesize (b) Masked Input}
        \put(3,62){\footnotesize (c) \mname(RGB)}
        \put(100,62){\footnotesize (d) \mname(RGB-D)}
        \put(-180,4){\footnotesize (e) BIPS$_{\text{\scriptsize{ECCV'2022}}}$}
        \put(-97,4){\footnotesize (f) OmniDreamer$_{\text{\scriptsize{CVPR'2022}}}$}
        \put(18,4){\footnotesize (g) TFill$_{\text{\scriptsize{CVPR'2022}}}$}
        \put(100,4){\footnotesize (h) Inpaint Anything$_{\text{\scriptsize{arXiv'2023}}}$}
    \end{picture}
    \vspace{-0.2cm}
    \caption{\textbf{Qualitative comparison for RGB panorama outpainting.} Our \mname generated more objects with appropriate layout, and with better visual quality. 
    For BIPS and OmniDreamer, despite the seemingly reasonable results, the outpainted areas tend to fill the walls and lack diverse items. 
    As for TFill, it generates blurry results for large invisible areas. 
    For Inpaint anything, it generates multiple objects but they appear to be structurally and semantically implausible.
    Compared to them, \mname generates more reasonable details in the masked region, such as pillows, paintings on the wall, windows, and views outside.
    More comparisons are provided in Appendix. 
    }
    \label{fig: RGB example}
\end{figure*}
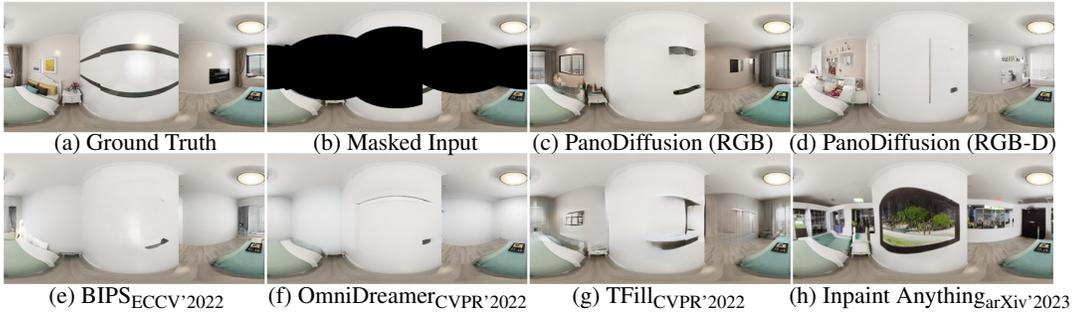

\begin{table}[tb!]
\begin{minipage}{\linewidth}{\begin{center}
    \renewcommand{\arraystretch}{1.0}
    \caption{\textbf{Quantitative results for RGB outpainting.} All models were re-trained and evaluated using the same standardized dataset. Note that, we tested all models \emph{without} the depth input.}
    \vspace{-0.2cm}
    \resizebox{\linewidth}{!}{
    \begin{tabular}{@{}l ccccc ccccc ccccc cccc@{}}
    \toprule
    \multirow{2}{*}{\textbf{Methods}} & \multicolumn{4}{c}{\textbf{Camera Mask}} && \multicolumn{4}{c}{\textbf{NFoV Mask}} && \multicolumn{4}{c}{\textbf{Layout Mask}} && \multicolumn{4}{c}{\textbf{Random Box Mask}} \\
    \cline{2-5}\cline{7-10}\cline{12-15}\cline{17-20}
    & FID $\downarrow$    & sFID $\downarrow$   & D $\uparrow$ & C $\uparrow$ && FID $\downarrow$    & sFID $\downarrow$   & D $\uparrow$ & C $\uparrow$ && FID $\downarrow$    & sFID $\downarrow$   & D $\uparrow$ & C $\uparrow$ && FID $\downarrow$    & sFID $\downarrow$   & D $\uparrow$ & C $\uparrow$ \\
    \midrule
    BIPS & 31.70 & 28.89 & 0.769 & 0.660 && 57.69 & 44.68 & 0.205 & 0.277 &   & 32.25 & 24.66 & 0.645 & 0.579 & & 25.35 & 22.60 & 0.676 & 0.798   \\
    OmniDreamer    & 65.47 & 37.04 & 0.143 & 0.175 & & 62.56 & 36.24 & 0.125 & 0.184 &  & 82.71 & 28.40 &  0.103  & 0.120 & & 45.10 & 24.12 & 0.329 & 0.576    \\
    LaMa    & 115.92 & 107.69 & 0.034 & 0.082 & & 125.77 & 136.32 & 0.002 & 0.006 &  & 129.77 & 35.23 &  0.018  & 0.043 & & 45.25 & 24.21 & 0.429 & 0.701    \\
    TFill    & 83.84 & 61.40 & 0.075 & 0.086 & & 93.62 & 76.13 & 0.037 & 0.027 & &  97.99 &  43.40  & 0.046 & 0.052 & & 46.84 & 30.72 & 0.368 & 0.574    \\
    Inpainting Anything    & 97.38 & 54.73 & 0.076 & 0.133 & & 105.77 & 59.70 & 0.054 & 0.035 & & 92.18  & 32.00   & 0.116 & 0.085 & & 46.30 & 26.71 & 0.372 &  0.632   \\
    RePaint    & 82.84 & 84.39 & 0.096 & 0.105 & & 95.38 & 82.35 & 0.0639 & 0.078 & & 69.14  &  31.63  & 0.294 & 0.263 & & 55.47 & 38.78 & 0.433 & 0.581    \\
    \mname     & \textbf{21.55} & \textbf{26.95} & \textbf{0.867} & \textbf{0.708} & & \textbf{21.41} & \textbf{27.80} & \textbf{0.790} & \textbf{0.669} &   & \textbf{23.06} & \textbf{22.39} & \textbf{1.000} & \textbf{0.737} &  & \textbf{16.13} & \textbf{20.39} & \textbf{1.000} & \textbf{0.883} \\
    \bottomrule
    \end{tabular}} 
    \label{tab:rgb_quant}
        \end{center}}
        \end{minipage}
    
\end{table}

Following prior works, we report the quantitative results for RGB panorama outpainting with camera masks in Table \ref{tab:rgb_quant}.
All instantiations of our model significantly outperform all state-of-the-art models.
Specifically, the FID score is substantially better (relative 67.0$\%$ improvement).

It is imperative to note that our model is trained unconditionally, with masks only employed during the inference phase. 
Therefore, it is expected to \emph{handle a broader spectrum of mask types}.
To validate this assertion, we further evaluated our model with the baseline models across all four different mask types (displayed in Fig.~\ref{fig: mask example}).
The results in Table \ref{tab:rgb_quant} show that \mname consistently outperforms the baseline models on all types of masks. 
Conversely, baseline models' performance displays significant variability in the type of mask used. 
Although the visible regions of the layout masks are always larger than the camera masks, the performances of baseline models on camera masks are significantly better.
This is likely because the masks in the training process are closer to the NFoV distribution. 
In contrast, \mname has a more robust performance, producing high-quality and diverse output images for all mask distributions.

The qualitative results are visualized in Fig.~\ref{fig: RGB example}. 
Here we show an example of outpainting on a layout mask (more comparisons in Appendix). 
Besides the fact that PanoDiffusion generates more visually realistic results than baseline models, comparing the RGB (trained without depth) and RGB-D versions of our \mname, in Fig.~\ref{fig: RGB example}(c), some unrealistic structures are generated on the center wall, and when we look closely at the curtains generated by the RGB model, the physical structure of the edges is not quite real. 
In contrast, the same region of
RGB-D result (Fig.~\ref{fig: RGB example}(d)) appears more structurally appropriate. 
Such improvement proves the advantages of jointly learning to synthesize depth data along with RGB images, \emph{even when depth is not used during test time}, suggesting the depth information is significant for assisting the RGB completion.

\subsection{Ablation Experiments}
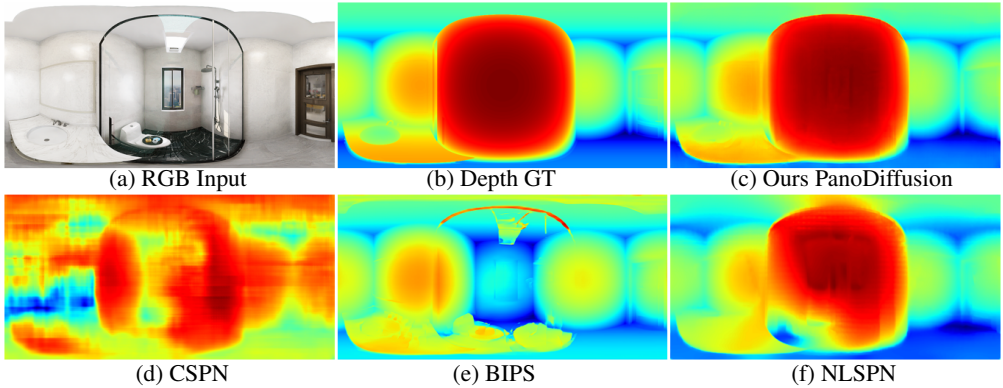
\begin{figure*}[tb!]
    \centering
    \includegraphics[width=0.95\linewidth]{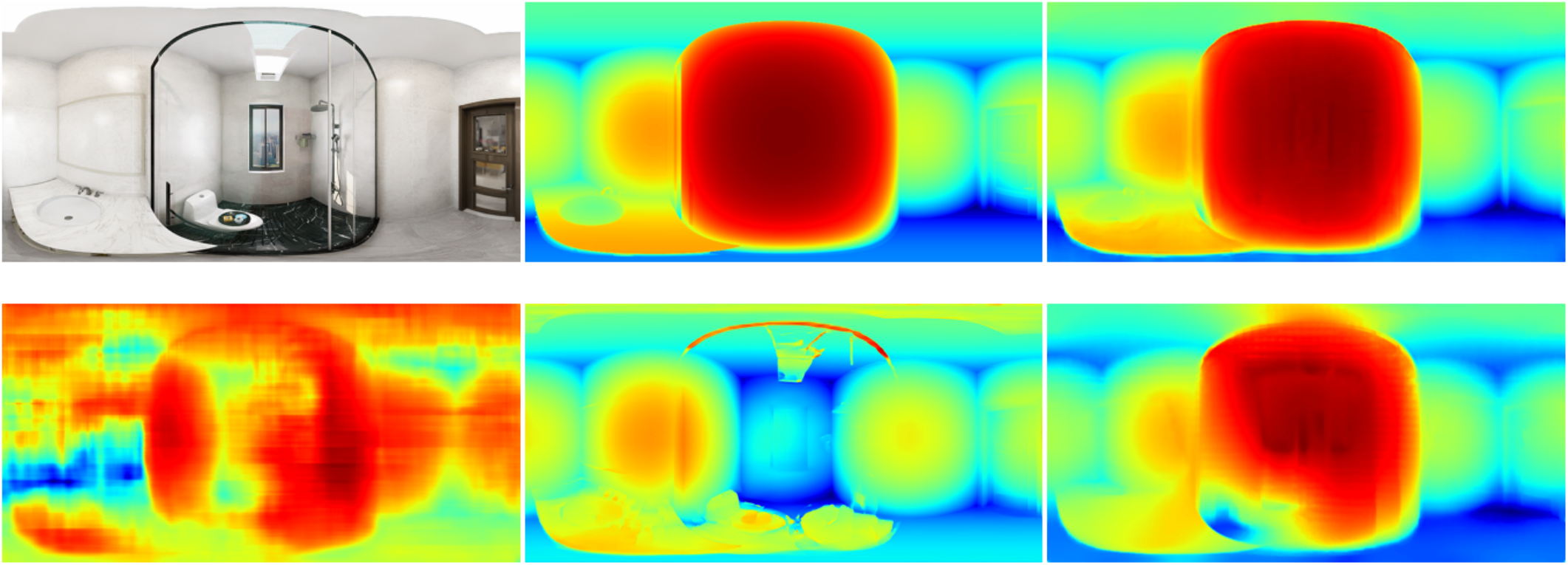}
    \begin{picture}(0,0)
        \put(-340,66){\footnotesize (a) RGB Input}
        \put(-220,66){\footnotesize (b) Depth GT}
        \put(-105,66){\footnotesize (c) Ours \mname}
        \put(-330,-8){\footnotesize (d) CSPN}
        \put(-210,-8){\footnotesize (e) BIPS}
        \put(-82,-8){\footnotesize (f) NLSPN}
    \end{picture}
    \vspace{4pt}
    \caption{\textbf{Qualitative comparison for depth panorama synthesis.}
    }
    \label{fig: Depth synthesis}
\end{figure*}

\begin{figure}[tb!]
    \begin{center}
            \centering
            \includegraphics[width=0.23\linewidth]{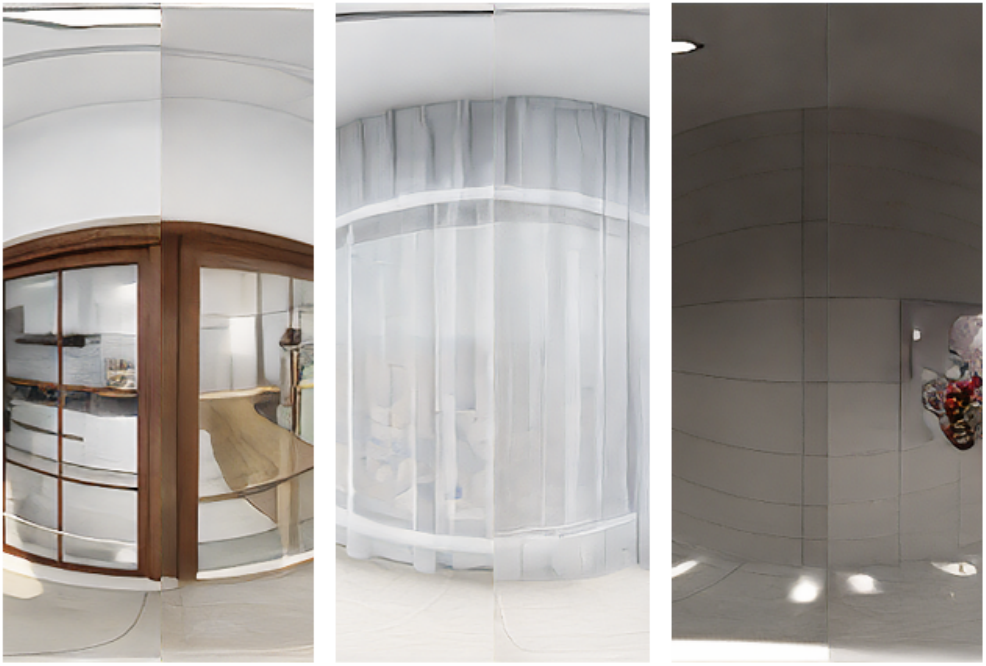}
        \hspace{0.1cm}
            \centering
            \includegraphics[width=0.23\linewidth]{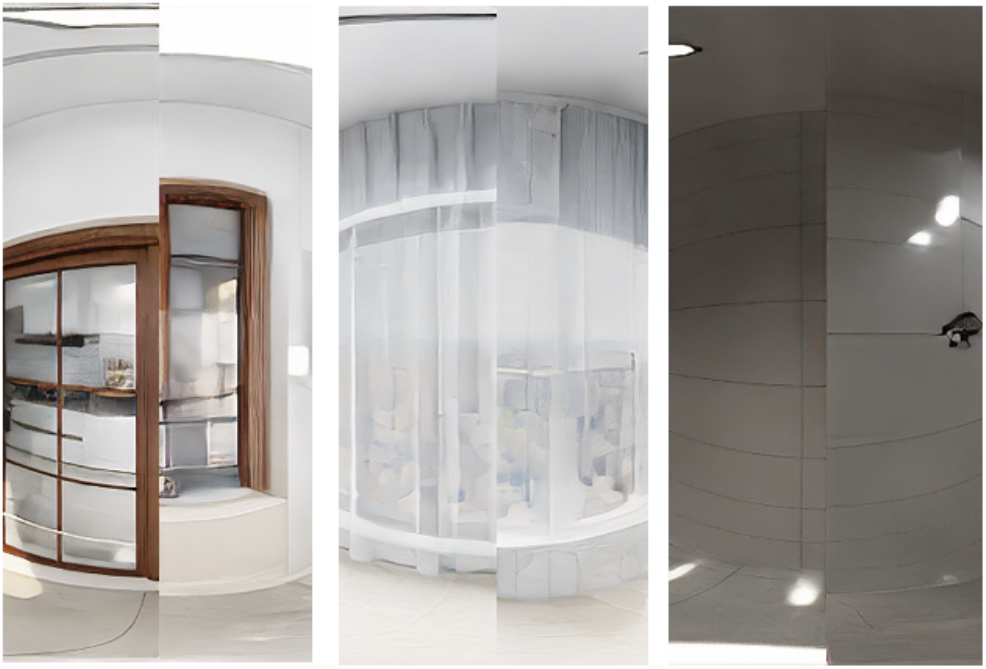}
        \hspace{0.1cm}
            \centering
            \includegraphics[width=0.23\linewidth]{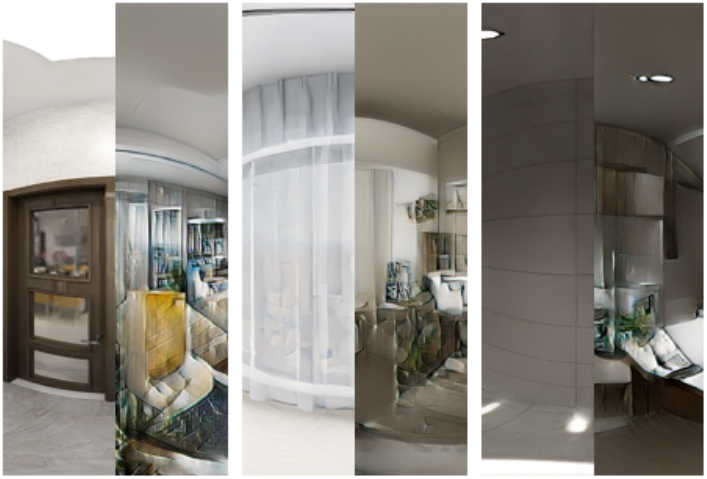}
        \hspace{0.1cm}
            \centering
            \includegraphics[width=0.23\linewidth]{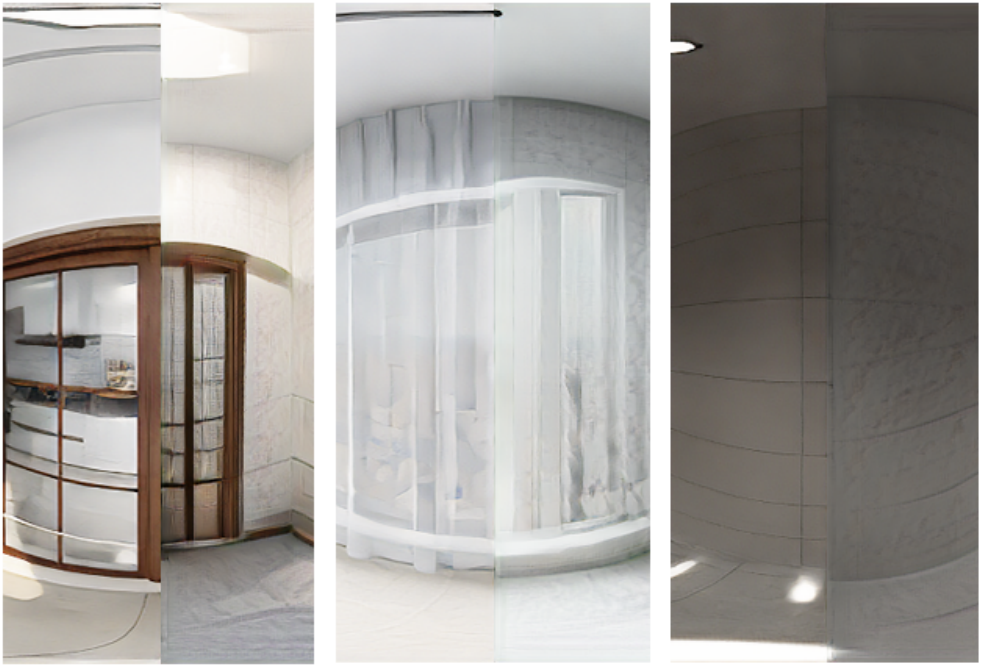}
        \begin{picture}(0,0)
            \put(-390,-8){\footnotesize (a) \mname w/ TA}
            \put(-293,-8){\footnotesize (b) \mname w/o TA}
            \put(-185,-8){\footnotesize (c) OmniDreamer}
            \put(-65,-8){\footnotesize (d) BIPS}
        \end{picture}
    \end{center}
    \caption{\textbf{Examples of stitched ends of the outpainted images.} For each image, the left half was unmasked (i.e. ground truth), while the right half was masked and synthesized. The results generated with rotation are more naturally connected at both ends (a).
    }
    \label{fig: two-end align examples} 
\end{figure}
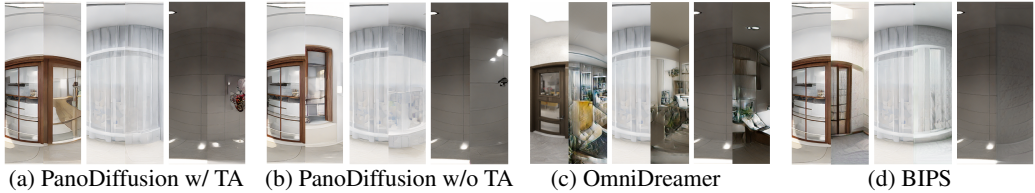

We ran a number of ablations to analyze the effectiveness of each core component in our \mname. Results are shown in \cref{tab:ablations,tab:LRCE} and 
\cref{fig: Depth synthesis,fig: two-end align examples} and discussed in detail next. 

\begin{table}[tb!]
    \begin{center}
    \caption{\textbf{Depth map ablations}. All models are trained and evaluated on the Structured3D dataset.}
    \label{tab:ablations}
    \vspace{-0.5cm}
    \subfigure{
    \begin{minipage}{\linewidth}{\begin{center}
    \renewcommand{\arraystretch}{1.0}
    \setlength\tabcolsep{3pt}
    \resizebox{\linewidth}{!}{
    \begin{tabular}{@{}l ccccc ccccc ccccc cccc@{}}
    \toprule
    \multirow{2}{*}{\textbf{Noise Level}} & \multicolumn{4}{c}{\textbf{Camera Mask}} && \multicolumn{4}{c}{\textbf{NFoV Mask}} && \multicolumn{4}{c}{\textbf{Layout Mask}} && \multicolumn{4}{c}{\textbf{Random Box Mask}} \\
    \cline{2-5}\cline{7-10}\cline{12-15}\cline{17-20}
    & FID $\downarrow$    & sFID $\downarrow$   & D $\uparrow$ & C $\uparrow$ && FID $\downarrow$    & sFID $\downarrow$   & D $\uparrow$ & C $\uparrow$ && FID $\downarrow$    & sFID $\downarrow$   & D $\uparrow$ & C $\uparrow$ && FID $\downarrow$    & sFID $\downarrow$   & D $\uparrow$ & C $\uparrow$ \\
    \midrule
    no depth     & 24.33 & 29.00 & 0.667 & 0.635 & & 24.01 & 30.00 & \cellcolor{cyan!20}0.639 & \cellcolor{magenta!20}0.617 &  & 25.37 & \cellcolor{blue!40}22.92 & 0.785 & 0.677 & & 17.88 & 21.21 & 0.913 & \cellcolor{magenta!60}0.857     \\
    50\%     & \cellcolor{orange!60}21.65 & \cellcolor{blue!20}28.12 &\cellcolor{cyan!20} 0.678 & \cellcolor{magenta!40}0.660 & & \cellcolor{orange!20}21.99 & \cellcolor{blue!60}29.37 & \cellcolor{cyan!60}0.678 & 0.561 &  & \cellcolor{orange!20}24.24 & 23.05 & \cellcolor{cyan!20}0.855 & \cellcolor{magenta!40}0.724 & & \cellcolor{orange!20}17.02 & \cellcolor{blue!40}21.22 & \cellcolor{cyan!20}0.919 &  0.837      \\
    30\%     & \cellcolor{orange!20}21.78 & \cellcolor{blue!40}27.96 &\cellcolor{cyan!40} 0.714 & \cellcolor{magenta!60}0.674 & & \cellcolor{orange!40}21.78 & \cellcolor{blue!40}29.39 & \cellcolor{cyan!40}0.643 & \cellcolor{magenta!60}0.658 &  & \cellcolor{orange!40}24.11 & \cellcolor{blue!20}23.00 & \cellcolor{cyan!40}0.919 & \cellcolor{magenta!40}0.724 &  & \cellcolor{orange!40}16.87 & \cellcolor{blue!20}21.25 & \cellcolor{cyan!60}0.937 & \cellcolor{magenta!40}0.855     \\
    10\%     & \cellcolor{orange!40}21.68 & \cellcolor{blue!60}27.79 &\cellcolor{cyan!60} 0.721 & \cellcolor{magenta!20}0.658 & & \cellcolor{orange!60}21.49 & \cellcolor{blue!20}29.74 & 0.558 & \cellcolor{magenta!40}0.620 &  & \cellcolor{orange!60}24.02 & \cellcolor{blue!60}22.68 & \cellcolor{cyan!60}0.938 & \cellcolor{magenta!70}\textbf{0.741} & & \cellcolor{orange!60}16.60 & \cellcolor{blue!60}21.02 &\cellcolor{cyan!40} 0.932 & \cellcolor{magenta!20}0.853\\
    full depth     & \cellcolor{orange!80}\textbf{21.55} & \cellcolor{blue!70}\textbf{26.95} & \cellcolor{cyan!70}\textbf{0.867} & \cellcolor{magenta!70}\textbf{0.708} & & \cellcolor{orange!80}\textbf{21.41} & \cellcolor{blue!70}\textbf{27.80} & \cellcolor{cyan!70}\textbf{0.790} & \cellcolor{magenta!70}\textbf{0.669} &   & \cellcolor{orange!80}\textbf{23.06} & \cellcolor{blue!70}\textbf{22.39} & \cellcolor{cyan!70}\textbf{1.000} & \cellcolor{magenta!60}0.737 &  & \cellcolor{orange!80}\textbf{16.13} & \cellcolor{blue!70}\textbf{20.39} & \cellcolor{cyan!70}\textbf{1.000} & \cellcolor{magenta!70}\textbf{0.883} \\
    \bottomrule
    \end{tabular}}  
        \end{center}}
        \footnotesize (a) \textbf{Usage of depth maps (training).} We use different sparsity levels of depth for training and the results (more intense color means better performance) verify the effectiveness of depth for RGB outpainting. It also proves that the model can accept sparse depth as input.   
        
        \end{minipage}
    }\\
    \subfigure{
    \begin{minipage}{0.45\linewidth}{\begin{center}
        \resizebox{\linewidth}{!}{
    \begin{tabular}{@{}lccccc@{}}
    \toprule
    \textbf{Methods} & Input Depth & FID $\downarrow$    & sFID $\downarrow$   & D $\uparrow$ & C $\uparrow$\\ 
    \midrule                
    
    BIPS       & \multirow{2}{*}{fully visible} & 29.74 & 30.59 &   \textbf{0.931}  &  \textbf{0.721}    \\
    \mname & & \textbf{21.90} & \textbf{26.78} &   0.829    &  0.693      \\
    \midrule
    BIPS       & \multirow{2}{*}{partial visible}& 31.70 & 28.89 & 0.769 & 0.660      \\
    \mname & & \textbf{22.34} & \textbf{26.74}  &  \textbf{0.856}    &  \textbf{0.686}      \\
    \midrule
    BIPS       & \multirow{2}{*}{fully masked}& 68.79 & 42.62 &  0.306   &  0.412   \\
    \mname & & \textbf{21.55} & \textbf{26.95} & \textbf{0.867} & \textbf{0.708}      \\
    \bottomrule
    \end{tabular}}
    \end{center}}
    \footnotesize (b) \textbf{Usage of depth maps (inference).} BIPS heavily relies on the availability of input depth during inference, while our model is minimally affected. 
    \end{minipage}
    \label{tab:test}
    }
    \subfigure{\begin{minipage}{0.515\linewidth}{\begin{center}
        \resizebox{\linewidth}{!}{
        
        \begin{tabular}{@{}lccccc@{}}
        \toprule
        \textbf{Methods} & Input Depth & RMSE $\downarrow$  & MAE $\downarrow$  & AbsREL $\downarrow$ & Delta1.25 $\uparrow$ \\ 
        \midrule                
        BIPS       &  \multirow{4}{*}{fully masked} & 323 & 207 &  0.1842   &  0.8436      \\
        CSPN       &   & 374 & 282 &  0.2273   &  0.6618      \\
        NLSPN &  &   284    &   \textbf{183}   &    0.1692     &   0.8544        \\
        \mname   &   &   \textbf{276}    &    193    &   \textbf{0.1355}      &    \textbf{0.9060}      \\
        \midrule
        BIPS         &  \multirow{4}{*}{partial visible}   & 247 & 136 &  0.1098   &  0.9032      \\
        CSPN  &  &   291    &  195    &   0.1547      &     0.8182      \\
        NLSPN &  &    221    &   124   &    \textbf{0.1058}     &  0.9143         \\
        \mname  &  &   \textbf{219}   &    \textbf{123}    &   0.1127      &     \textbf{0.9278}      \\
        \bottomrule
        \end{tabular}}
    \end{center}}
    \footnotesize (c) \textbf{Depth panorama synthesis.} Our model outperforms baseline models in most of the metrics.
    \end{minipage}
    } 
    \end{center}
    \vspace{-0.5cm}
\end{table}

\vspace{-12pt}
\paragraph{Depth Maps.} 
In practice applications, depth data may exhibit sparsity due to the hardware limitations \citep{park2020non}.
To ascertain the model's proficiency in accommodating sparse depth maps as input,
we undertook a \textbf{training} process using depth maps with different degrees of sparsity (i.e., randomized depth value will be set to 0). 
The result is reported in Table~\ref{tab:ablations}(a). 
The denser colors in the table represent better performance. 
As the sparsity of the depth input decreases, the performance of RGB outpainting constantly improves. 
Even if we use 50\% sparse depth for training, the result is overall better than the original LDM. 

We then evaluated the importance of depth maps during \textbf{inference}, and compared it with the state-of-the-art BIPS \citep{oh2022bips}, which is also trained with RGB-D images.
The quantitative results are reported in Table~\ref{tab:ablations}(b). 
As can be seen, BIPS's performance appears to deteriorate significantly when the input depth visual area is reduced. 
Conversely, our \mname is \emph{not sensitive to these depth maps}, indicating that the generic model has successfully handled the modality. 
Interestingly, we noticed that having fully visible depth at test time did \emph{not} improve the performance of \mname, and in fact, the result deteriorated slightly. 
A reasonable explanation is that during the training process, the signal-to-noise ratios (SNR) of RGB and depth pixels are roughly the same within each iteration since no masks were used. 
However, during outpainting, the SNR balance will be disrupted when RGB input is masked and depth input is fully visible. 
Therefore, the results are degraded, but only slightly because \mname has effectively learned the distribution of spatial visual patterns across all modalities, without being overly reliant on depth. 
This also explains why our model is more robust to depth inputs with different degrees of visibility. 

Finally, we evaluated the depth synthesis ability of \mname, seen in Table~\ref{tab:ablations}(c) and Fig.~\ref{fig: Depth synthesis}. 
 The results show that our model achieves the best performance on most of the metrics and the qualitative results also show that \mname is able to accurately estimate the depth map. 
This not only indicates that \mname can be used for depth synthesis and estimation but also proves that it has learned the spatial patterns of panorama images.

\vspace{-8pt}
\paragraph{Two-end Alignment.} 
Currently, there is no metric to evaluate the performance of aligning the two ends of an image. 
To make a reasonable comparison, we make one side of the input image fully visible, and the other side fully masked and then compare the two ends of output.
Based on the Left-Right Consistency Error (LRCE) \citep{shen2022panoformer} which is used to evaluate the consistency of two ends of the depth maps, we designed a new RGB-LRCE to calculate the difference between the two ends of the image: $LRCE=\frac{1}{N}\sum_{i=1}^N|P_{first}^{col}-P_{last}^{col}|$, and reported results in \cref{tab:LRCE}.

\begin{table}[tb!]
\caption{\textbf{Two-end alignment ablations.} Using rotational outpainting, we achieve optimal consistency at both ends of the \mname output.}
\label{tab:LRCE}
\vspace{-6pt}
\resizebox{\linewidth}{!}{
\begin{tabular}{@{}lcccc|lcccc@{}}
\toprule
\textbf{Methods $\backslash$ Mask Type} & Camera    & NfoV   & Layout & Random Box & \textbf{Methods $\backslash$ Mask Type} & Camera    & NfoV   & Layout & Random Box\\ 
\midrule                
\mname(w/ rotation) &   \textbf{90.41}    &    \textbf{89.74}    &   \textbf{88.01}      &     \textbf{85.04} &  \mname(w/o rotation)  & 125.82 & 128.33 & 128.10 & 128.19    \\
BIPS   & 117.59 & 96.82 &  132.15   &  148.78 & OmniDreamer        & 115.6 & 109.00 &    146.37    &    136.68      \\ 
LaMa  & 119.51 & 119.39 & 133.54 & 136.35 & TFill & 155.16 & 157.60 & 136.94 & 122.96 \\
\bottomrule
\end{tabular}}
\end{table}

The qualitative results are shown in ~\cref{fig: two-end align examples}. 
To compare as many results, we only show the end regions that are stitched together to highlight the contrast. 
They show that the consistency of the two ends of the results is improved after the use of rotational outpainting, especially the texture of the walls and the alignment of the layout. 
Still, differences can be found with rotated outpainting. 
We believe it is mainly due to the fact that rotational denoising is based on the latent level, which may introduce extra errors during decoding.

\section{Conclusion}
In this paper, we show that our proposed method, the two-stage RGB-D \mname, achieves state-of-the-art performance for indoor RGB-D panorama outpainting. 
The introduction of depth information via our bi-modal LDM structure significantly improves the performance of the model. 
Such improvement illustrates the effectiveness of using depth during training as an aid to guide RGB panorama generation. 
In addition, we show that the alignment mechanism we employ at each step of the denoising process of the diffusion model enhances the wraparound consistency of the results. 
With the use of these novel mechanisms, our two-stage structure is capable of generating high-quality RGB-D panoramas at 512$\times$1024 resolution.

\bibliography{iclr2024_conference}

\begin{thebibliography}{41}
\providecommand{\natexlab}[1]{#1}
\providecommand{\url}[1]{\texttt{#1}}
\expandafter\ifx\csname urlstyle\endcsname\relax
  \providecommand{\doi}[1]{doi: #1}\else
  \providecommand{\doi}{doi: \begingroup \urlstyle{rm}\Url}\fi

\bibitem[Akimoto et~al.(2019)Akimoto, Kasai, Hayashi, and Aoki]{akimoto2019360}
Naofumi Akimoto, Seito Kasai, Masaki Hayashi, and Yoshimitsu Aoki.
\newblock 360-degree image completion by two-stage conditional gans.
\newblock In \emph{2019 IEEE International Conference on Image Processing
  (ICIP)}, pp.\  4704--4708. IEEE, 2019.

\bibitem[Akimoto et~al.(2022)Akimoto, Matsuo, and Aoki]{akimoto2022diverse}
Naofumi Akimoto, Yuhi Matsuo, and Yoshimitsu Aoki.
\newblock Diverse plausible 360-degree image outpainting for efficient 3dcg
  background creation.
\newblock In \emph{Proceedings of the IEEE/CVF Conference on Computer Vision
  and Pattern Recognition}, pp.\  11441--11450, 2022.

\bibitem[Cheng et~al.(2018)Cheng, Wang, and Yang]{cheng2018depth}
Xinjing Cheng, Peng Wang, and Ruigang Yang.
\newblock Depth estimation via affinity learned with convolutional spatial
  propagation network.
\newblock In \emph{Proceedings of the European conference on computer vision
  (ECCV)}, pp.\  103--119, 2018.

\bibitem[Choi et~al.(2015)Choi, Zhou, and Koltun]{choi2015robust}
Sungjoon Choi, Qian-Yi Zhou, and Vladlen Koltun.
\newblock Robust reconstruction of indoor scenes.
\newblock In \emph{Proceedings of the IEEE Conference on Computer Vision and
  Pattern Recognition}, pp.\  5556--5565, 2015.

\bibitem[Epstein et~al.(2022)Epstein, Park, Zhang, Shechtman, and
  Efros]{epstein2022blobgan}
Dave Epstein, Taesung Park, Richard Zhang, Eli Shechtman, and Alexei~A Efros.
\newblock Blobgan: Spatially disentangled scene representations.
\newblock In \emph{Computer Vision--ECCV 2022: 17th European Conference, Tel
  Aviv, Israel, October 23--27, 2022, Proceedings, Part XV}, pp.\  616--635.
  Springer, 2022.

\bibitem[Gardner et~al.(2017)Gardner, Sunkavalli, Yumer, Shen, Gambaretto,
  Gagn{\'e}, and Lalonde]{gardner2017learning}
Marc-Andr{\'e} Gardner, Kalyan Sunkavalli, Ersin Yumer, Xiaohui Shen, Emiliano
  Gambaretto, Christian Gagn{\'e}, and Jean-Fran{\c{c}}ois Lalonde.
\newblock Learning to predict indoor illumination from a single image.
\newblock \emph{ACM Transactions on Graphics (TOG)}, 36\penalty0 (6):\penalty0
  1--14, 2017.

\bibitem[Gardner et~al.(2019)Gardner, Hold-Geoffroy, Sunkavalli, Gagn{\'e}, and
  Lalonde]{gardner2019deep}
Marc-Andr{\'e} Gardner, Yannick Hold-Geoffroy, Kalyan Sunkavalli, Christian
  Gagn{\'e}, and Jean-Fran{\c{c}}ois Lalonde.
\newblock Deep parametric indoor lighting estimation.
\newblock In \emph{Proceedings of the IEEE/CVF International Conference on
  Computer Vision}, pp.\  7175--7183, 2019.

\bibitem[Goodfellow et~al.(2014)Goodfellow, Pouget-Abadie, Mirza, Xu,
  Warde-Farley, Ozair, Courville, and Bengio]{goodfellow2014generative}
Ian Goodfellow, Jean Pouget-Abadie, Mehdi Mirza, Bing Xu, David Warde-Farley,
  Sherjil Ozair, Aaron Courville, and Yoshua Bengio.
\newblock Generative adversarial nets.
\newblock \emph{Advances in neural information processing systems}, 27, 2014.

\bibitem[Hara et~al.(2021)Hara, Mukuta, and Harada]{hara2021spherical}
Takayuki Hara, Yusuke Mukuta, and Tatsuya Harada.
\newblock Spherical image generation from a single image by considering scene
  symmetry.
\newblock In \emph{Proceedings of the AAAI Conference on Artificial
  Intelligence}, volume~35, pp.\  1513--1521, 2021.

\bibitem[Heusel et~al.(2017)Heusel, Ramsauer, Unterthiner, Nessler, and
  Hochreiter]{heusel2017gans}
Martin Heusel, Hubert Ramsauer, Thomas Unterthiner, Bernhard Nessler, and Sepp
  Hochreiter.
\newblock Gans trained by a two time-scale update rule converge to a local nash
  equilibrium.
\newblock \emph{Advances in neural information processing systems}, 30, 2017.

\bibitem[Ho et~al.(2020)Ho, Jain, and Abbeel]{ho2020denoising}
Jonathan Ho, Ajay Jain, and Pieter Abbeel.
\newblock Denoising diffusion probabilistic models.
\newblock \emph{Advances in Neural Information Processing Systems},
  33:\penalty0 6840--6851, 2020.

\bibitem[Horita et~al.(2022)Horita, Yang, Chen, Koyama, and
  Aizawa]{horita2022structure}
Daichi Horita, Jiaolong Yang, Dong Chen, Yuki Koyama, and Kiyoharu Aizawa.
\newblock A structure-guided diffusion model for large-hole diverse image
  completion.
\newblock \emph{arXiv preprint arXiv:2211.10437}, 2022.

\bibitem[Iizuka et~al.(2017)Iizuka, Simo-Serra, and
  Ishikawa]{iizuka2017globally}
Satoshi Iizuka, Edgar Simo-Serra, and Hiroshi Ishikawa.
\newblock Globally and locally consistent image completion.
\newblock \emph{ACM Transactions on Graphics (ToG)}, 36\penalty0 (4):\penalty0
  1--14, 2017.

\bibitem[Kingma \& Welling(2014)Kingma and Welling]{kingma2013auto}
Diederik~P Kingma and Max Welling.
\newblock Auto-encoding variational bayes.
\newblock In editor (ed.), \emph{Proceedings of the International Conference on
  Learning Representations (ICLR)}, 2014.

\bibitem[Lee et~al.(2017)Lee, Yea, Park, and Yoon]{lee2017joint}
Jeong-Kyun Lee, Jaewon Yea, Min-Gyu Park, and Kuk-Jin Yoon.
\newblock Joint layout estimation and global multi-view registration for indoor
  reconstruction.
\newblock In \emph{Proceedings of the IEEE international conference on computer
  vision}, pp.\  162--171, 2017.

\bibitem[Li et~al.(2022)Li, Yu, Zhou, Song, Lin, and Jia]{li2022sdm}
Wenbo Li, Xin Yu, Kun Zhou, Yibing Song, Zhe Lin, and Jiaya Jia.
\newblock Sdm: Spatial diffusion model for large hole image inpainting.
\newblock \emph{arXiv preprint arXiv:2212.02963}, 2022.

\bibitem[Lu et~al.(2023)Lu, Hu, Wang, Bai, and Wang]{lu2023autoregressive}
Zhuqiang Lu, Kun Hu, Chaoyue Wang, Lei Bai, and Zhiyong Wang.
\newblock Autoregressive omni-aware outpainting for open-vocabulary 360-degree
  image generation.
\newblock \emph{arXiv preprint arXiv:2309.03467}, 2023.

\bibitem[Lugmayr et~al.(2022)Lugmayr, Danelljan, Romero, Yu, Timofte, and
  Van~Gool]{lugmayr2022repaint}
Andreas Lugmayr, Martin Danelljan, Andres Romero, Fisher Yu, Radu Timofte, and
  Luc Van~Gool.
\newblock Repaint: Inpainting using denoising diffusion probabilistic models.
\newblock In \emph{Proceedings of the IEEE/CVF Conference on Computer Vision
  and Pattern Recognition}, pp.\  11461--11471, 2022.

\bibitem[Naeem et~al.(2020)Naeem, Oh, Uh, Choi, and Yoo]{naeem2020reliable}
Muhammad~Ferjad Naeem, Seong~Joon Oh, Youngjung Uh, Yunjey Choi, and Jaejun
  Yoo.
\newblock Reliable fidelity and diversity metrics for generative models.
\newblock In \emph{International Conference on Machine Learning}, pp.\
  7176--7185. PMLR, 2020.

\bibitem[Nash et~al.(2021)Nash, Menick, Dieleman, and
  Battaglia]{nash2021generating}
Charlie Nash, Jacob Menick, Sander Dieleman, and Peter~W Battaglia.
\newblock Generating images with sparse representations.
\newblock \emph{arXiv preprint arXiv:2103.03841}, 2021.

\bibitem[Newcombe et~al.(2011)Newcombe, Izadi, Hilliges, Molyneaux, Kim,
  Davison, Kohi, Shotton, Hodges, and Fitzgibbon]{newcombe2011kinectfusion}
Richard~A Newcombe, Shahram Izadi, Otmar Hilliges, David Molyneaux, David Kim,
  Andrew~J Davison, Pushmeet Kohi, Jamie Shotton, Steve Hodges, and Andrew
  Fitzgibbon.
\newblock Kinectfusion: Real-time dense surface mapping and tracking.
\newblock In \emph{2011 10th IEEE international symposium on mixed and
  augmented reality}, pp.\  127--136. Ieee, 2011.

\bibitem[Oh et~al.(2022)Oh, Cho, Chae, Park, Wang, and Yoon]{oh2022bips}
Changgyoon Oh, Wonjune Cho, Yujeong Chae, Daehee Park, Lin Wang, and Kuk-Jin
  Yoon.
\newblock Bips: Bi-modal indoor panorama synthesis via residual depth-aided
  adversarial learning.
\newblock In \emph{Computer Vision--ECCV 2022: 17th European Conference, Tel
  Aviv, Israel, October 23--27, 2022, Proceedings, Part XVI}, pp.\  352--371.
  Springer, 2022.

\bibitem[Park et~al.(2020)Park, Joo, Hu, Liu, and So~Kweon]{park2020non}
Jinsun Park, Kyungdon Joo, Zhe Hu, Chi-Kuei Liu, and In~So~Kweon.
\newblock Non-local spatial propagation network for depth completion.
\newblock In \emph{Computer Vision--ECCV 2020: 16th European Conference,
  Glasgow, UK, August 23--28, 2020, Proceedings, Part XIII 16}, pp.\  120--136.
  Springer, 2020.

\bibitem[Pathak et~al.(2016)Pathak, Krahenbuhl, Donahue, Darrell, and
  Efros]{pathak2016context}
Deepak Pathak, Philipp Krahenbuhl, Jeff Donahue, Trevor Darrell, and Alexei~A
  Efros.
\newblock Context encoders: Feature learning by inpainting.
\newblock In \emph{Proceedings of the IEEE conference on computer vision and
  pattern recognition}, pp.\  2536--2544, 2016.

\bibitem[Ren et~al.(2012)Ren, Bo, and Fox]{ren2012rgb}
Xiaofeng Ren, Liefeng Bo, and Dieter Fox.
\newblock Rgb-(d) scene labeling: Features and algorithms.
\newblock In \emph{2012 IEEE Conference on Computer Vision and Pattern
  Recognition}, pp.\  2759--2766. IEEE, 2012.

\bibitem[Rombach et~al.(2022)Rombach, Blattmann, Lorenz, Esser, and
  Ommer]{rombach2022high}
Robin Rombach, Andreas Blattmann, Dominik Lorenz, Patrick Esser, and Bj{\"o}rn
  Ommer.
\newblock High-resolution image synthesis with latent diffusion models.
\newblock In \emph{Proceedings of the IEEE/CVF Conference on Computer Vision
  and Pattern Recognition}, pp.\  10684--10695, 2022.

\bibitem[Salimans et~al.(2016)Salimans, Goodfellow, Zaremba, Cheung, Radford,
  and Chen]{salimans2016improved}
Tim Salimans, Ian Goodfellow, Wojciech Zaremba, Vicki Cheung, Alec Radford, and
  Xi~Chen.
\newblock Improved techniques for training gans.
\newblock \emph{Advances in neural information processing systems}, 29, 2016.

\bibitem[Shen et~al.(2022)Shen, Lin, Liao, Nie, Zheng, and
  Zhao]{shen2022panoformer}
Zhijie Shen, Chunyu Lin, Kang Liao, Lang Nie, Zishuo Zheng, and Yao Zhao.
\newblock Panoformer: Panorama transformer for indoor 360-degree depth
  estimation.
\newblock In \emph{European Conference on Computer Vision}, pp.\  195--211.
  Springer, 2022.

\bibitem[Somanath \& Kurz(2021)Somanath and Kurz]{somanath2021hdr}
Gowri Somanath and Daniel Kurz.
\newblock Hdr environment map estimation for real-time augmented reality.
\newblock In \emph{Proceedings of the IEEE/CVF Conference on Computer Vision
  and Pattern Recognition}, pp.\  11298--11306, 2021.

\bibitem[Song \& Funkhouser(2019)Song and Funkhouser]{song2019neural}
Shuran Song and Thomas Funkhouser.
\newblock Neural illumination: Lighting prediction for indoor environments.
\newblock In \emph{Proceedings of the IEEE/CVF Conference on Computer Vision
  and Pattern Recognition}, pp.\  6918--6926, 2019.

\bibitem[Suvorov et~al.(2022)Suvorov, Logacheva, Mashikhin, Remizova, Ashukha,
  Silvestrov, Kong, Goka, Park, and Lempitsky]{suvorov2022resolution}
Roman Suvorov, Elizaveta Logacheva, Anton Mashikhin, Anastasia Remizova,
  Arsenii Ashukha, Aleksei Silvestrov, Naejin Kong, Harshith Goka, Kiwoong
  Park, and Victor Lempitsky.
\newblock Resolution-robust large mask inpainting with fourier convolutions.
\newblock In \emph{Proceedings of the IEEE/CVF winter conference on
  applications of computer vision}, pp.\  2149--2159, 2022.

\bibitem[Tang et~al.(2023)Tang, Zhang, Chen, Wang, and
  Furukawa]{tang2023MVDiffusion}
Shitao Tang, Fuyang Zhang, Jiacheng Chen, Peng Wang, and Yasutaka Furukawa.
\newblock Mvdiffusion: Enabling holistic multi-view image generation with
  correspondence-aware diffusion.
\newblock \emph{arXiv preprint 2307.01097}, 2023.

\bibitem[Wan et~al.(2021)Wan, Zhang, Chen, and Liao]{Wan_2021_ICCV}
Ziyu Wan, Jingbo Zhang, Dongdong Chen, and Jing Liao.
\newblock High-fidelity pluralistic image completion with transformers.
\newblock In \emph{Proceedings of the IEEE/CVF International Conference on
  Computer Vision (ICCV)}, pp.\  4692--4701, October 2021.

\bibitem[Wang et~al.(2023)Wang, Saharia, Montgomery, Pont-Tuset, Noy,
  Pellegrini, Onoe, Laszlo, Fleet, Soricut, et~al.]{wang2023imagen}
Su~Wang, Chitwan Saharia, Ceslee Montgomery, Jordi Pont-Tuset, Shai Noy,
  Stefano Pellegrini, Yasumasa Onoe, Sarah Laszlo, David~J Fleet, Radu Soricut,
  et~al.
\newblock Imagen editor and editbench: Advancing and evaluating text-guided
  image inpainting.
\newblock In \emph{Proceedings of the IEEE/CVF Conference on Computer Vision
  and Pattern Recognition}, pp.\  18359--18369, 2023.

\bibitem[Xie et~al.(2023)Xie, Zhang, Lin, Hinz, and Zhang]{xie2023smartbrush}
Shaoan Xie, Zhifei Zhang, Zhe Lin, Tobias Hinz, and Kun Zhang.
\newblock Smartbrush: Text and shape guided object inpainting with diffusion
  model.
\newblock In \emph{Proceedings of the IEEE/CVF Conference on Computer Vision
  and Pattern Recognition}, pp.\  22428--22437, 2023.

\bibitem[Yu et~al.(2018)Yu, Lin, Yang, Shen, Lu, and Huang]{yu2018generative}
Jiahui Yu, Zhe Lin, Jimei Yang, Xiaohui Shen, Xin Lu, and Thomas~S Huang.
\newblock Generative image inpainting with contextual attention.
\newblock In \emph{Proceedings of the IEEE Conference on Computer Vision and
  Pattern Recognition}, pp.\  5505--5514, 2018.

\bibitem[Yu et~al.(2023)Yu, Feng, Feng, Liu, Jin, Zeng, and
  Chen]{yu2023inpaint}
Tao Yu, Runseng Feng, Ruoyu Feng, Jinming Liu, Xin Jin, Wenjun Zeng, and Zhibo
  Chen.
\newblock Inpaint anything: Segment anything meets image inpainting.
\newblock \emph{arXiv preprint arXiv:2304.06790}, 2023.

\bibitem[Zhao et~al.(2020)Zhao, Cui, Sheng, Dong, Liang, Eric, Chang, and
  Xu]{zhao2020large}
Shengyu Zhao, Jonathan Cui, Yilun Sheng, Yue Dong, Xiao Liang, I~Eric, Chao
  Chang, and Yan Xu.
\newblock Large scale image completion via co-modulated generative adversarial
  networks.
\newblock In \emph{International Conference on Learning Representations}, 2020.

\bibitem[Zheng et~al.(2019)Zheng, Cham, and Cai]{zheng2019pluralistic}
Chuanxia Zheng, Tat-Jen Cham, and Jianfei Cai.
\newblock Pluralistic image completion.
\newblock In \emph{Proceedings of the IEEE/CVF Conference on Computer Vision
  and Pattern Recognition}, pp.\  1438--1447, 2019.

\bibitem[Zheng et~al.(2021)Zheng, Cham, and Cai]{zheng2021pluralistic}
Chuanxia Zheng, Tat-Jen Cham, and Jianfei Cai.
\newblock Pluralistic free-form image completion.
\newblock In \emph{International Journal of Computer Vision}, pp.\  2786--2805, 2021.

\bibitem[Zheng et~al.(2019)Zheng, Cham, and Cai]{zheng2018t2net}
Chuanxia Zheng, Tat-Jen Cham, and Jianfei Cai.
\newblock T2net: Synthetic-to-realistic translation for solving single-image depth estimation tasks.
\newblock In \emph{Proceedings of the European conference on computer vision (ECCV)}, pp.\  767--783, 2018.

\bibitem[Zheng et~al.(2022)Zheng, Cham, Cai, and Phung]{zheng2022bridging}
Chuanxia Zheng, Tat-Jen Cham, Jianfei Cai, and Dinh Phung.
\newblock Bridging global context interactions for high-fidelity image
  completion.
\newblock In \emph{Proceedings of the IEEE/CVF Conference on Computer Vision
  and Pattern Recognition}, pp.\  11512--11522, 2022.

\bibitem[Zheng et~al.(2020)Zheng, Zhang, Li, Tang, Gao, and
  Zhou]{zheng2020structured3d}
Jia Zheng, Junfei Zhang, Jing Li, Rui Tang, Shenghua Gao, and Zihan Zhou.
\newblock Structured3d: A large photo-realistic dataset for structured 3d
  modeling.
\newblock In \emph{Computer Vision--ECCV 2020: 16th European Conference,
  Glasgow, UK, August 23--28, 2020, Proceedings, Part IX 16}, pp.\  519--535.
  Springer, 2020.

\end{thebibliography}
\bibliographystyle{iclr2024_conference}

\end{document}